\documentclass{article}
\usepackage[final, nonatbib]{neurips_2021}
\usepackage{times}
\usepackage{epsfig}
\usepackage{amssymb}
\usepackage{multirow}
\usepackage{wrapfig}
\usepackage{float}
\usepackage[utf8]{inputenc} 
\usepackage[T1]{fontenc}    
\usepackage{url}            
\usepackage{booktabs}       
\usepackage{amsfonts}       
\usepackage{nicefrac}       
\usepackage{microtype}      
\usepackage{xcolor}         
\usepackage{xspace}

\usepackage{hyperref}
\usepackage{graphicx}
\usepackage{amsmath}
\usepackage{listings}
\usepackage{multirow}
\usepackage{todonotes}
\usepackage{placeins}

\usepackage{threeparttable}

\title{EPSANet: An Efficient Pyramid Squeeze Attention Block on Convolutional Neural Network}

\author{%
	Hu Zhang$^{1}$ \ \ \ \ \ Keke Zu$^{1}$ \ \ \ \ \ Jian Lu$^{1}$\thanks{Corresponding author.} \ \ \ \ \ Yuru Zou$^{1}$ \ \ \ \ \ Deyu Meng$^{2}$ \\
	$^1$ Shenzhen University, Shenzhen, China \  \   \ $^2$Xi’an Jiaotong University, Xi'an, China \\
	{huzhang198}@gmail.com \  \{kekezu, jianlu, yuruzou\}@szu.edu.cn \ {dymeng}@mail.xjtu.edu.cn\\
}

\begin{document}
\maketitle

\begin{abstract}
  Recently, it has been demonstrated that the performance of a deep convolutional neural network can be effectively improved by embedding an attention module into it. In this work, a novel lightweight and effective attention method named Pyramid Squeeze Attention (PSA) module is proposed. By replacing the 3x3 convolution with the PSA module in the bottleneck blocks of the ResNet, a novel representational block named Efficient Pyramid Squeeze Attention (EPSA) is obtained. The EPSA block can be easily added as a plug-and-play component into a well-established backbone network, and significant improvements on model performance can be achieved. Hence, a simple and efficient backbone architecture named EPSANet is developed in this work by stacking these ResNet-style EPSA blocks. Correspondingly, a stronger multi-scale representation ability can be offered by the proposed EPSANet for various computer vision tasks including but not limited to, image classification, object detection, instance segmentation, etc. Without bells and whistles, the performance of the proposed EPSANet outperforms most of the state-of-the-art channel attention methods. As compared to the SENet-50, the Top-1 accuracy is improved by 1.93$\%$ on ImageNet dataset, a larger margin of +2.7 box AP for object detection and an improvement of +1.7 mask AP for instance segmentation by using the Mask-RCNN on MS-COCO dataset are obtained. Our source code is available at:\href{https://github.com/murufeng/EPSANet}{https://github.com/murufeng/EPSANet}. 
\end{abstract}

\section{Introduction} 
Attention mechanisms are widely used in many computer vision areas such as image classification, object detection, instance segmentation, semantic segmentation, scene parsing and action localization~\cite{2019GCNet,deeplab,ASFF,resatt,cbam,Zhao2017PSP,SANet_seg}. Specifically, there are two types of attention methods, which are channel attention and spatial attention. Recently, it has been demonstrated that significant performance improvements can be achieved by employing the channel attention, spatial attention, or both of them~\cite{doubleAAnet,2020FcaNet,2020PCANet,20ecanet,zhang2020resnest}. The most commonly used method of channel attention is the Squeeze-and-Excitation (SE) module \cite{hu2018senet}, which can significantly improve the performance with a considerably low cost. The drawback of the SENet is that it ignores the importance of spatial information. Therefore, the Bottleneck Attention Module(BAM)~\cite{BAM} and Convolutional Block Attention Module(CBAM) \cite{cbam} are proposed to enrich the attention map by effectively combining the spatial and channel attention. However, there still exists two important and challenging problems. The first one is how to efficiently capture and exploit the spatial information of the feature map with different scales to enrich the feature space. The second one is that the channel or spatial attention can only effectively capture the local information but fail in establishing a long-range channel dependency. Correspondingly, many methods are proposed to address these two problems. The methods based on multi-scale feature representation and cross-channel information interaction, such as the PyConv \cite{pyconv}, the Res2Net \cite{res2net}, and the HS-ResNet \cite{hs-resnet}, are proposed. In the other hand, a long-range channel dependency can be established as shown in \cite{deeplab,DAN_dualatt,Non-local_long}. All the above mentioned methods, however, bring higher model complexity and thus the network suffers from heavy computational burden.
\begin{figure}
	\centering
	\includegraphics[width=0.8\linewidth]{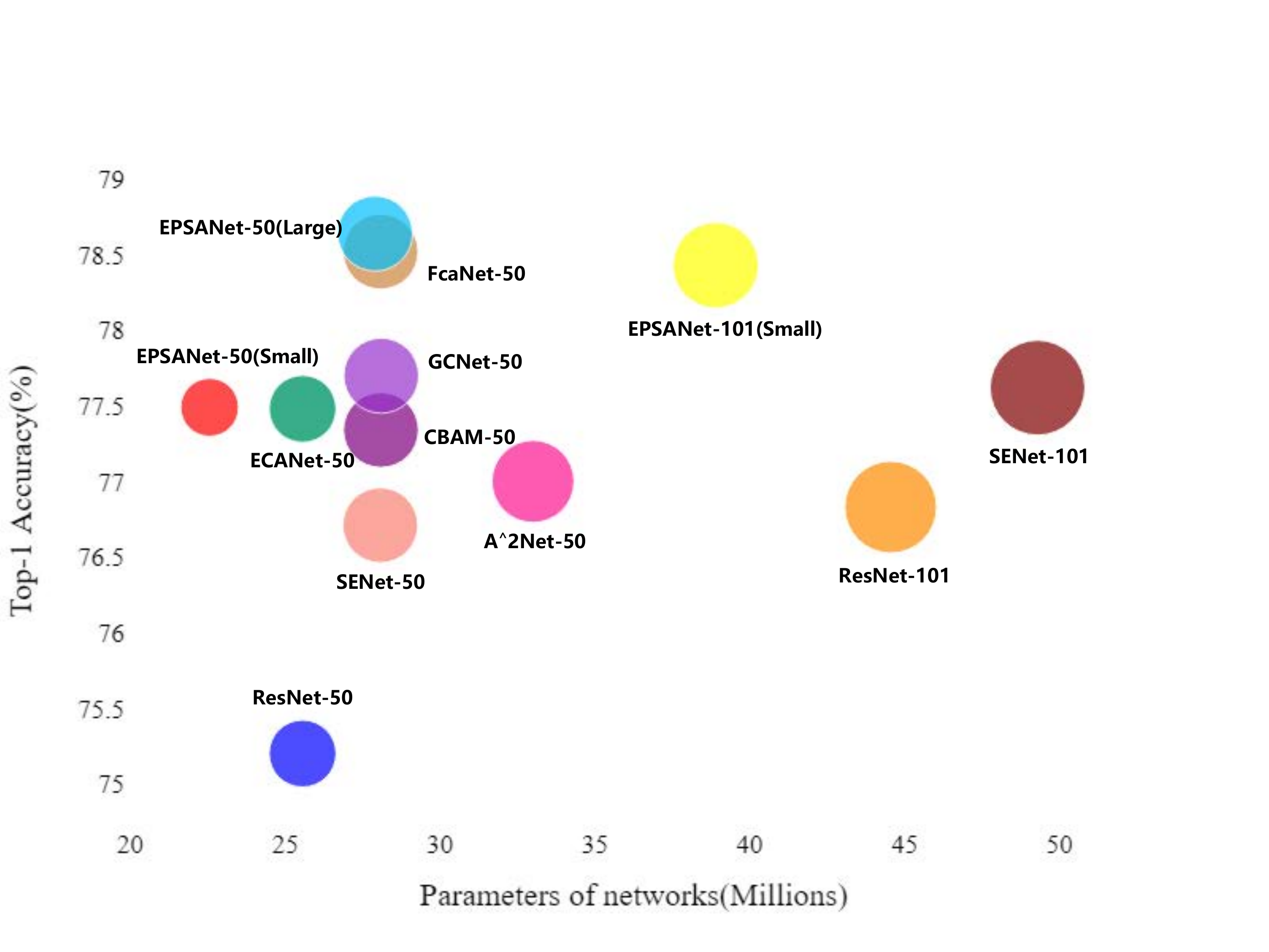}
	\caption{Comparing the accuracy of different 
		attention methods with ResNet-50 and ResNet-101 as our backbone. The circle area reflects the network parameters and FLOPs of different models.} 
	\label{fig:epsanet-parameters}
\end{figure}
Based on the above observations, we see it is necessary to develop a low-cost but effective attention module. In this work, a low-cost and high-performance novel module named Pyramid Squeeze Attention (PSA) is proposed. The proposed PSA module has the ability to process the input tensor at multiple scales. Specifically, by using the multi-scale pyramid convolution structure to integrate the information of the input feature map. Meanwhile, we can effectively extract the spatial information with different scales from each channel-wise feature map by squeezing the channel dimension of the input tensor. By doing this, neighbor scales of context features can be more precisely incorporated. Finally, a cross-dimension interaction is built by extracting the channel-wise attention weight of the multi-scale feature maps. The Softmax operation is employed to recalibrate the attention weight of the corresponding channels, and thus a long-range channel dependency is established. As a result, a novel block named Efficient Pyramid Squeeze Attention (EPSA) is obtained by replacing the 3x3 convolution with the PSA module in the bottleneck blocks of the ResNet. Furthermore, a network named EPSANet is proposed by stacking these EPSA blocks as the ResNet style. As shown by Figure \ref{fig:epsanet-parameters}, the proposed EPSANet can not only outperform prior arts in terms of the Top-1 accuracy but is also more efficient in terms of required parameters. The main contributions of this work are summarized as below:

\begin{itemize}
	\item A novel Efficient Pyramid Squeeze Attention (EPSA) block is proposed, which can effectively extract multi-scale spatial information at a more granular level and develop a long-range channel dependency. The proposed EPSA block is very flexible and scalable and thus can be applied to a large variety of network architectures for numerous tasks of computer vision.
	
	\item A novel backbone architecture named EPSANet is proposed, which can learn richer multi-scale feature representation and adaptively re-calibrate the cross-dimension channel-wise attention weight. 
	
	\item Extensive experiments demonstrated that promising results can be achieved by the proposed EPSANet across image classification, object detection and instance segmentation on both ImageNet and COCO datasets.
\end{itemize}

\section{Related Work}
\textbf{Attention mechanism}  The attention mechanism is used to strength the allocation of the most informative feature expressions while suppressing the less useful ones, and thus makes the model attending to important regions within a context adaptively. The Squeeze-and-Excitation (SE) attention in \cite{hu2018senet} can capture channel correlations by selectively modulating the scale of channel. The CBAM in \cite{cbam} can enrich the attention map by adding max pooled features for the channel attention with large-size kernels. Motivated by the CBAM, the GSoP in \cite{GSoP} proposed a second-order pooling method to extract richer feature aggregation. More recently, the Non-Local block \cite{Non-local_long} is proposed to build a dense spatial feature map and capture the long-range dependency via non-local operations. Based on the Non-Local block, the Double Attention Network($A^2$Net) \cite{doubleAAnet} introduces a novel relation function to embed the attention with spatial information into the feature map. Sequently, the SKNet in \cite{sknet} introduces a dynamic selection attention mechanism that allows each neuron to adaptively adjust its receptive field size based on multiple scales of input feature map. The ResNeSt \cite{zhang2020resnest} proposes a similar Split-Attention block that enables attention across groups of the input feature map. The Fcanet \cite{2020FcaNet} proposes a novel multi-spectral channel attention that realizes the pre-processing of channel attention mechanism in the frequency domain. The GCNet \cite{2019GCNet} introduces a simple spatial attention module and thus a long-range channel dependency is developed. The ECANet \cite{20ecanet} employs the one-dimensional convolution layer to reduce the redundancy of fully connected layers. The DANet \cite{DAN_dualatt} adaptively integrates local features with their global dependencies by summing these two attention modules from different branches. The above mentioned methods either focus on the design of more sophisticated attention modules that inevitably bring a greater computational cost, or they cannot establish a long-range channel dependency. Thus, in order to further improve the efficiency and reduce the model complexity, a novel attention module named PSA is proposed, which aims at learning attention weight with low model complexity and to effectively integrate local and global attention for establishing the long-range channel dependency.

\textbf{Multi-scale Feature Representations}  The ability of the multi-scale feature representation is essential for various vision tasks such as, instance segmentation \cite{he2017mask}, face anaxlysis \cite{face}, object detection \cite{ren2015faster}, salient object detection \cite{salient_object_pyramid}, and semantic segmentation \cite{SANet_seg}. It is critically important to design a good operator that can extract multi-scale feature more efficiently for visual recognition tasks. By embedding a operator for  multi-scale feature extraction into a convolution neural network(CNN), a more effective feature representation ability can be obtained. In the other hand, CNNs can naturally learn coarse-to-fine multi-scale features through a stack of convolutional operators. Thus, to design a better convolutional operator is the key for improving the multi-scale representations of CNNs. 

\section{Method}

\subsection{Revisting Channel Attention}
\begin{wrapfigure}{r}{9cm}
	\centering
	\includegraphics[width=0.8\linewidth]{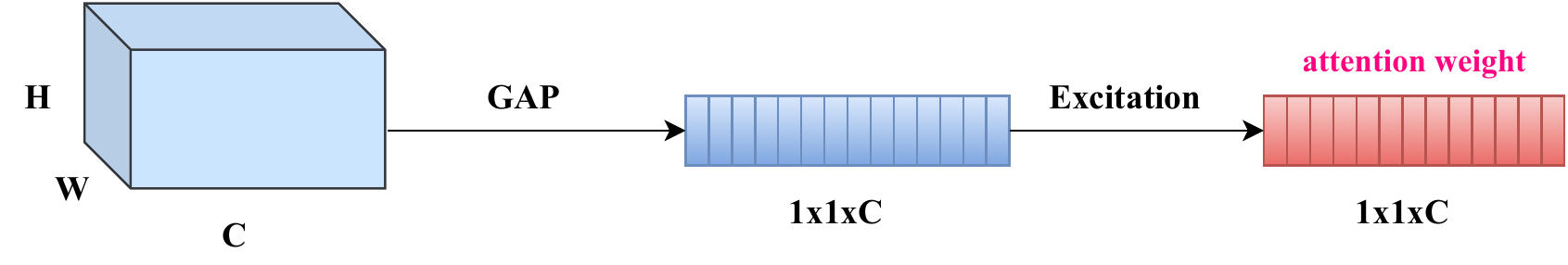}
	\caption{SEWeight module.}
	\label{fig:se_w_module}
\end{wrapfigure}
\textbf{Channel attention}  The channel attention mechanism allows the network to selectively weight the importance of each channel and thus generates more informative outputs. Let $X\in \mathbb{R}^{C\times H\times W}$ denotes the input feature map, where the quantity $H$, $W$, $C$ represent its height, width, number of input channels respectively. A SE block consists of two parts: squeeze and excitation, which is respectively designed for encoding the global information and adaptively recalibrating the channel-wise relationship. Generally, the channel-wise statistics can be generated by using a global average pooling, which is used to embed the global spatial information into a channel descriptor. The global average pooling operator can be calculated by the following equation
\begin{equation}
	g_{c} = \frac{1}{H \times W} \sum\limits_{i=1}^H \sum\limits_{j=1}^W x_{c}(i,j)
\end{equation}
The attention weight of the $c$-th channel in the SE block can be written as
\begin{equation}
	w_{c} = \sigma(W_{1} \delta (W_{0}(g_{c})))
\end{equation}
where the symbol $\delta $ represents the Recitified Linear Unit (ReLU) operation as in \cite{relu}, $W_{0}\in \mathbb{R}^{C\times\frac{C}{r}}$ and $W_{1}\in \mathbb{R}^{\frac{C}{r} \times C}$ represent the fully-connected (FC) layers. With two fully-connected layers, the linear information among channels can be combined more efficiently, and it is helpful for the interaction of the information of high and low channel dimensions. The symbol $\sigma$ represents the excitation function, and a Sigmoid function is usually used in practice. By using the excitation function, we can assign weights to channels after the channel interaction and thus the information can be extracted more efficiently. The above introduced process of generating channel attention weights is named SEWeight module in \cite{hu2018senet}, the diagram of the SEWeight module is shown by Figure \ref{fig:se_w_module}. 

\subsection{PSA Module}
\begin{figure}[H]
	\centering
	\includegraphics[width=0.8\linewidth]{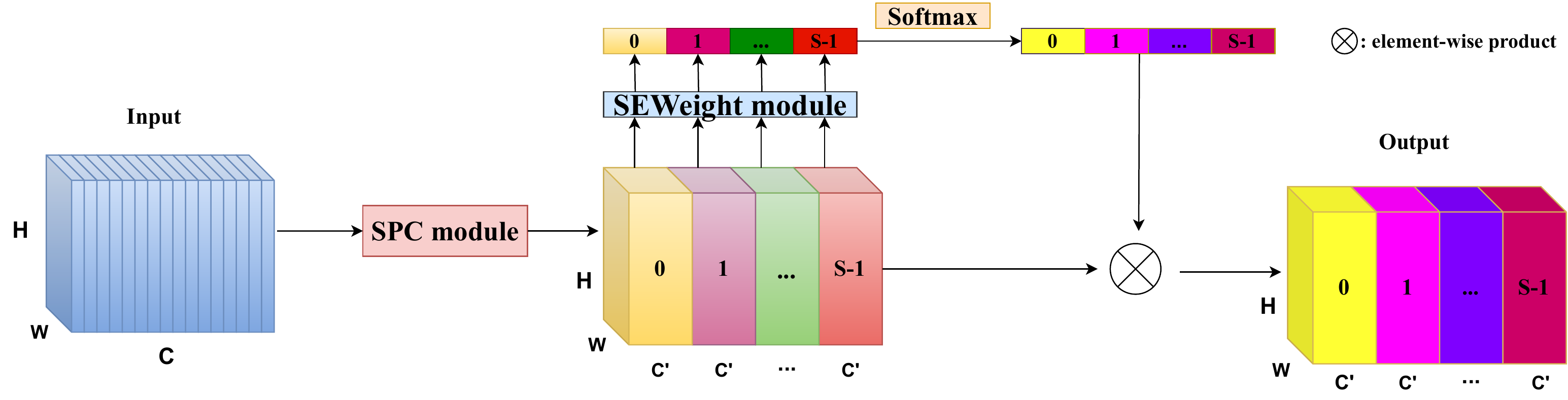}
	\caption{The structure of the proposed Pyramid Squeeze Attention (PSA) module.}
	\label{fig:psa_module}
\end{figure}

The motivation of this work is to build a more efficient and effective channel attention mechanism. Therefore, a novel pyramid squeeze attention (PSA) module is proposed. As illustrated by Figure \ref{fig:psa_module}, the PSA module is mainly implemented in four steps. First, the multi-scale feature map on channel-wise is obtained by implementing the proposed Squeeze and Concat (SPC) module. Second, the channel-wise attention vector are obtained by using the SEWeight module to extract the attention of the feature map with different scales. Third, re-calibrated weight of multi-scale channel is obtained by using the Softmax to re-calibrate the channel-wise attention vector. Fourth, the operation of an element-wise product is applied to the re-calibrated weight and the corresponding feature map. Finally, a refined feature map which is richer in multi-scale feature information can be obtained as the output.

\begin{figure}[H]
	\centering
	\includegraphics[width=0.8\linewidth]{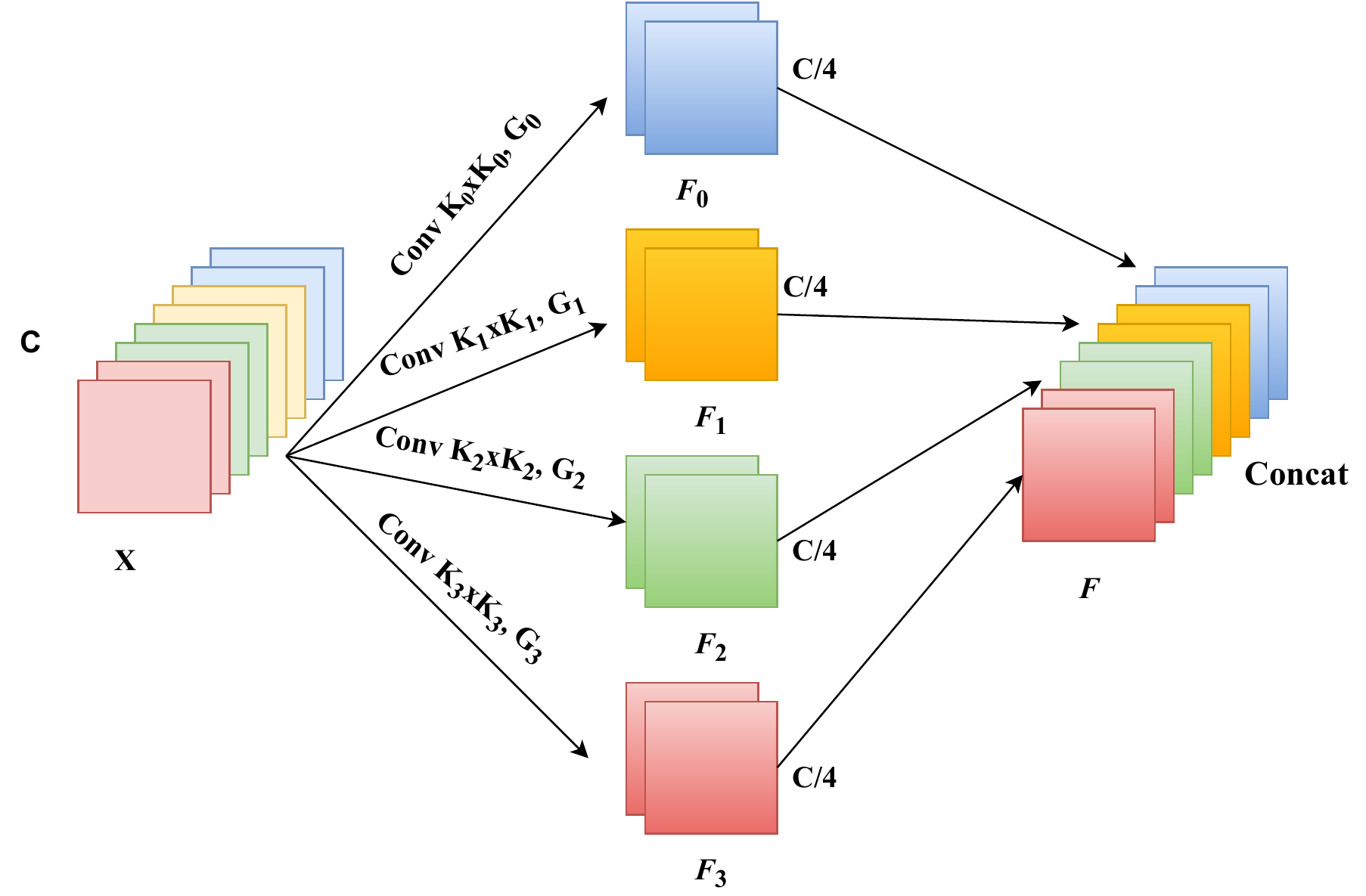}
	\caption{A detailed illustration of the proposed Squeeze and Concat (SPC) module with $S$=4, where 'Squeeze' means to equally squeeze in the channel dimension, $K$ is the kernel size, $G$ is the group size and 'Concat' means to concatenate features in the channel dimension.}
	\label{fig:spc_module}
\end{figure}

As illustrated by Figure \ref{fig:spc_module}, the essential operator for implementing the multi-scale feature extraction in the proposed PSA is the SPC, we extract the spatial information of the input feature map in a multi-branch way, the input channel dimension of each branch is $C$. By doing this, we can obtain more abundant positional information of the input tensor and process it at multiple scales in a parallel way. Thus a feature map that contains a single type of kernel can be obtained. Correspondingly, the different spatial resolutions and depths can be generated by using multi-scale convolutional kernels in a pyramid structure. And the spatial information with different scales on each channel-wise feature map can be effectively extracted by squeezing the channel dimension of the input tensor. Finally, each featur map with different scales $F_{i}$ has the common channel dimension $C^{'}=\frac{C}{S}$ and $i = {0,1,\cdots,S-1}$. Note that $C$ should be divisible by $S$. For each branch, it learns the multi-scale spatial information independently and establishs a cross-channel interaction in a local manner. However, a huge improvement in the amount of parameters will be resulted with the increase of kernel sizes. In order to process the input tensor at different kernel scales without increasing the computational cost, a method of group convolution is introduced and applied to the convolutional kernels. Further, we design a novel criterion for choosing the group size without increasing the number of parameters. The relationship between the multi-scale kernel size and the group size can be written as 
\begin{equation}
	G = 2^{\frac{K-1}{2}}
\end{equation}
where the quantity $K$ is the kernel size, $G$ is the group size. The above equation has been verifed by our ablation experiments, especially when $k\times k $ is equal to $3 \times 3$ and the default of G is 1. Finally, the multi-scale feature map generation function is given by 
\begin{equation}
	F_{i} = {\rm Conv}(k_{i} \times k_{i}, G_{i})(X) \quad i=0,1,2 \cdots S-1
\end{equation}
 where the $i$-th kernel size $k_{i} = 2 \times (i+1) + 1$, the $i$-th group size $G_{i} = 2 ^{\frac{k_{i}-1}{2}}$ and $F_{i} \in  R^{C^{'} \times H \times W}$ denotes the feature map with different scales. The whole multi-scale pre-processed feature map can be obtained by a concatenation way as
\begin{equation}
	F = {\rm Cat}([F_{0}, F_{1}, \cdots, F_{S-1}])	
\end{equation}
where $F \in R^{C \times H \times W}$ is the obtained multi-scale feature map. By extracting the channel attention weight information from the multi-scale pre-processed feature map, the attention weight vectors with different scales are obtained. Mathematically, the vector of attention weight can be represented as
\begin{equation}
	Z_{i} = {\rm SEWeight}(F_{i}), \quad i=0,1,2 \cdots S-1
\end{equation} where $Z_{i} \in R^{C^{'} \times 1 \times 1}$ is the attention weight. The SEWeight module is used to obtain the attention weight from the input feature map with different scales. By doing this, our PSA module can fuse context information in different scales and produce a better pixel-level attention for high-level feature maps. Further, in order to realize the interaction of attention information and fuse the cross-dimensions vector without destroying the orginal channel attention vector. And thus the whole multi-scale channel attention vector is obtained in a concatenation way as
\begin{equation}
	Z= Z_{0}\oplus Z_{1} \oplus \cdots \oplus Z_{S-1}
\end{equation}
where $\oplus$ is the concat operator, $Z_{i}$ is the attention value from the $F_{i}$, $Z$ is the multi-scale attention weight vector. A soft attention is used across channels to adaptively select different spatial scales, which is guided by the compact feature descriptor $Z_{i}$. A soft assignment weight is given by
\begin{equation}
	att_{i}= {\rm Softmax}(Z_{i})= \frac{exp(Z_{i})}{\sum_{i=0}^{S-1}exp(Z_{i})}
\end{equation}
where the Softmax is used to obtain the re-calibrated weight $att_{i}$ of the multi-scale channel, which contains all the location information on the space and the attention weight in channel. By doing this, the interaction between local and global channel attention is realized. Next, the channel attention of feature re-calibration is fused and spliced in a concatenation way, and thus the whole channel attention vector can be obtained as
\begin{equation}
	att = att_{0} \oplus att_{1} \oplus \cdots \oplus att_{S-1}
\end{equation}
where $att$ represents the multi-scale channel weight after attention interaction. Then, we multiply the re-calibrated weight of multi-scale channel attention $att_{i}$ with the feature map of the corresponding scale $F_{i}$ as
\begin{equation}
	Y_{i} = F_{i} \odot att_{i} \quad i=1, 2,3,\cdots S-1
\end{equation}
where $\odot$ represents the channel-wise multiplication, $Y_{i}$ refers to the feature map that with the obtained multi-scale channel-wise attention weight. The concatenation operator is more effective than the summation due to it can integrally maintain the feature representation without destroying the information of the orginal feature map. In sum, the process to obtain the refined output can be written as
\begin{equation}
	Out = {\rm Cat}([Y_{0},Y_{1},\cdots, Y_{S-1}])
\end{equation}
As illustrated by the above analysis, our proposed PSA module can integrate the multi-scale spatial information and the cross-channel attention into the block for each feature group. Thus, a better information interaction between local and global channel attention can be obtained by our proposed PSA module. 
\begin{figure}
	\centering
	\includegraphics[width=0.8\linewidth]{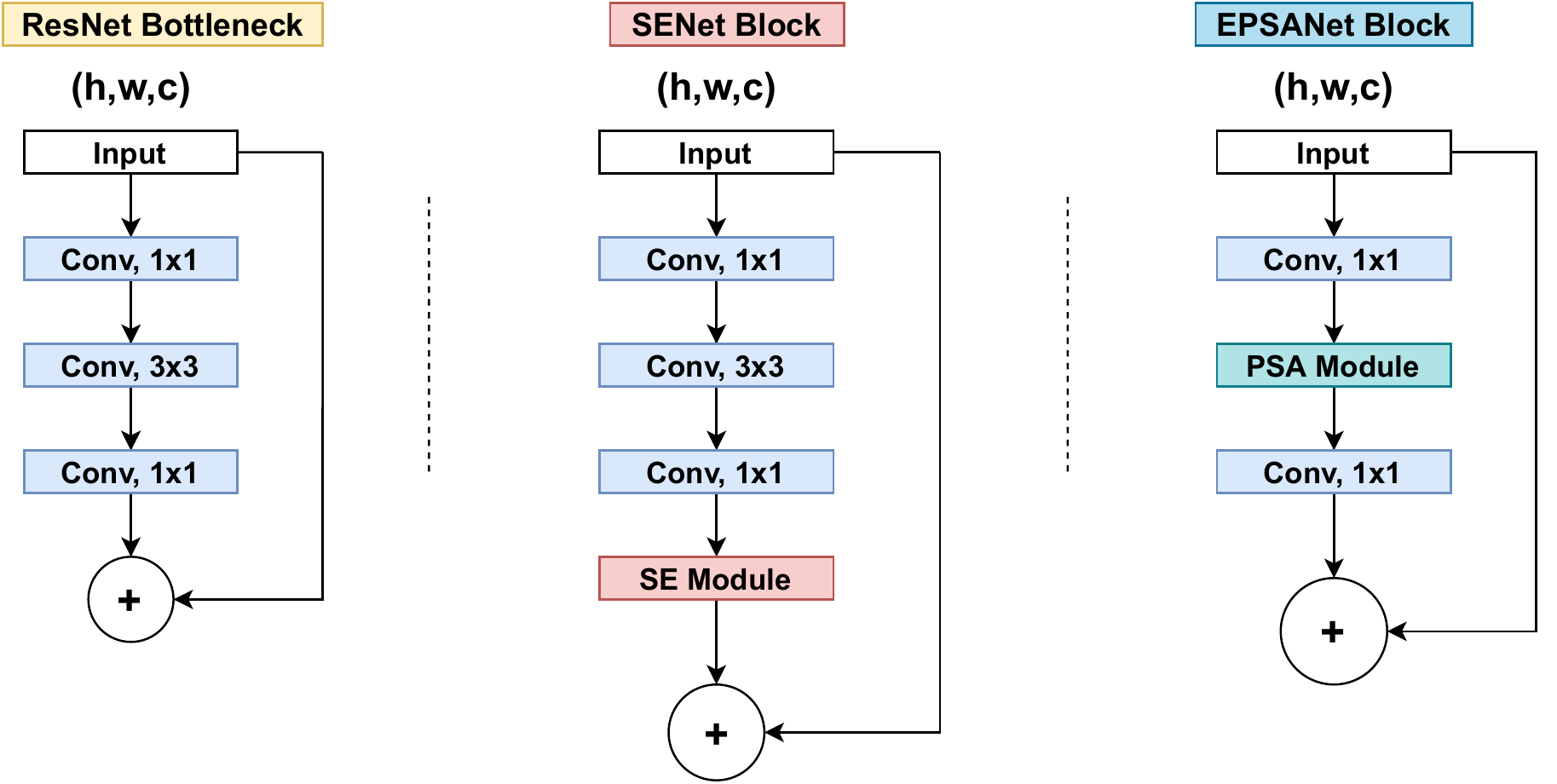}
	\caption{Illustration and comparison of ResNet, SENet, and our proposed EPSANet blocks.} 
	\label{fig:epsanet-block}
\end{figure}
\begin{table}[H]
	\centering
	\caption{Network design of the proposed EPSANet. }
	\begin{tabular}{cccc} \hline
		Output      & ResNet-50        & EPSANet-50(Small)     & EPSANet-50(Large)  \\ \hline
		112$\times$112  & \multicolumn{3}{c}{7$\times$7, 64, stride 2} \\
		\hline
		56$\times$56      & \multicolumn{3}{c}{3$\times$3 max pool, stride 2} \\ \hline
		56$\times$56      
		& 
		$\begin{bmatrix}
			1\times1, & 64 \\
			3\times3, & 64 \\
			1\times1, & 256 \\
		\end{bmatrix}$
		$\times3$
		
		& 
		$\begin{bmatrix}
			1\times1, & 64 \\
			PSA, & 64 \\
			1\times1, & 256 \\
		\end{bmatrix}$
		$\times3$
		
		& 
		$\begin{bmatrix}
			1\times1, & 128 \\
			PSA(G=32), & 128 \\
			1\times1, & 256 \\
		\end{bmatrix}$
		$\times3$
		\\  \hline
		
		28$\times$28     
		& 
		$\begin{bmatrix}
			1\times1, & 128 \\
			3\times3, & 128 \\
			1\times1, & 512 \\
		\end{bmatrix}$
		$\times4$
		
		& 
		$\begin{bmatrix}
			1\times1, & 128 \\
			PSA, & 128 \\
			1\times1, & 512 \\
		\end{bmatrix}$
		$\times4$
		
		& 
		$\begin{bmatrix}
			1\times1, & 256 \\
			PSA(G=32), & 256 \\
			1\times1, & 512 \\
		\end{bmatrix}$
		$\times4$	
		\\
		\hline
		
		14$\times$14     
		& 
		$\begin{bmatrix}
			1\times1, & 256 \\
			3\times3, & 256 \\
			1\times1, & 1024 \\
		\end{bmatrix}$
		$\times6$
		& 
		$\begin{bmatrix}
			1\times1, & 256 \\
			PSA, & 256 \\
			1\times1, & 1024 \\
		\end{bmatrix}$
		$\times6$
		
		& 
		$\begin{bmatrix}
			1\times1, & 512 \\
			PSA(G=32), & 512 \\
			1\times1, & 1024 \\
		\end{bmatrix}$
		$\times6$
		\\
		\hline
		
		7$\times$7     
		& 
		$\begin{bmatrix}
			1\times1, & 512 \\
			3\times3, & 512 \\
			1\times1, & 2048 \\
		\end{bmatrix}$
		$\times3$
		& 
		$\begin{bmatrix}
			1\times1, & 512 \\
			PSA, & 512 \\
			1\times1, & 2048 \\
		\end{bmatrix}$
		$\times3$
		
		& 
		$\begin{bmatrix}
			1\times1, & 1024 \\
			PSA(G=32), & 1024 \\
			1\times1, & 2048 \\
		\end{bmatrix}$
		$\times3$
		
		\\
		\hline
		1 $\times$ 1 &  \multicolumn{3}{c}{7$\times$7 global average pool,1000-d fc} \\ \hline
		
	\end{tabular}
	\label{table:network}
\end{table}
\subsection{Network Design}
 As shown by Figure \ref{fig:epsanet-block}, a novel block named Efficient Pyramid Squeeze Attention (EPSA) block is further obtained by replacing the 3x3 convolution with the PSA module at corresponding positions in the bottelneck blocks of ResNet. The multi-scale spatial information and the cross-channel attention are integrated by our PSA module into the EPSA block. Thus, the EPSA block can extract multi-scale spatial information at a more granular level and develop a long-range channel dependency. Correspondingly, a novel efficient backbone network named EPSANet is developed by stacking the proposed EPSA blocks as the ResNet style. The proposed EPSANet inherits the advantages of the EPSA block, and thus it has strong multi-scale representation capabilities and can adaptively re-calibrate the cross-dimension channel-wise weight. As shown by Table \ref{table:network}, two variations of the EPSANet, the EPSANet(Small) and EPSANet(Large) are proposed. For the proposed EPSANet(Small), the kernel and group size are respectively set as (3,5,7,9) and (1,4,8,16) in the SPC module. The proposed EPSANet(Large) has a higher group size and is set as (32,32,32,32) in the SPC module. 

\begin{table}[H]\centering \label{table:1}
	\caption{ Comparison of various attention methods on ImageNet in terms of network parameters(in millions), floating point operations per second (FLOPs), Top-1 and Top-5 Validation Accuracy($\%$).}
	\begin{tabular}{|l|l|c|c|c|c|}\hline
		Network         & Backbone  &Parameters  & FLOPs & Top-1 Acc. ($\%$)&Top-5 Acc. ($\%$)\\ \hline
		ResNet\cite{he2016deep}          & \multirow{9}{18mm}{ResNet-50}  & {25.56} &4.12G & 75.20  & 92.52    \\ 
		SENet\cite{hu2018senet}           & & 28.07  & 4.13G    & 76.71          & 93.38             \\ 
		CBAM\cite{cbam}            & & 28.07  &4.14G    & 77.34          & 93.69             \\
		A$^{2}$-Net\cite{doubleAAnet}      & & 33.00  &6.50G    & 77.00          & 93.50             \\
		ABN\cite{ABN}            & & 43.59  &7.18G    & 76.90          & -                  \\
		GCNet\cite{2019GCNet}          & & 28.11  &4.13G    & 77.70          & 93.66               \\ 
		Triplet Attention\cite{triplet_attention} & &25.56 &4.17G &77.48 & 93.68 \\
		AANet\cite{AAnet}  & & 25.80  &4.15G    & 77.70          & 93.80               \\
		ECANet\cite{20ecanet}         & & 25.56  &4.13G    & 77.48          & 93.68       \\  
		FcaNet\cite{2020FcaNet}        &  & 28.07  &4.13G  & 78.52   & 94.14 \\	
		EPSANet(Small)     & & \textbf{22.56}  & \textbf{3.62G}   & 77.49  & 93.54     \\ 
		EPSANet(Large)  & & 27.90  & 4.72G & \textbf{78.64} & \textbf{94.18}  \\
		
		\hline
		ResNet\cite{he2016deep}         & \multirow{11}{18mm}{ResNet-101}  & {44.55} &7.85G  & 76.83 & 93.48    \\ 
		SENet\cite{hu2018senet}         &  & 49.29           &7.86G     & 77.62        &93.93         \\ 
		BAM\cite{BAM}            &  & 44.91           &7.93G     & 77.56           & 93.71    \\ 
		CBAM\cite{cbam}           &  & 49.33           &7.88G      & 78.49       & 94.31             \\
		SRM\cite{SRM}  &  &44.68            &7.95G     & 78.47        & 94.20         \\
		ECANet\cite{20ecanet}         &  & 44.55           &7.86G     & 78.65           & 94.34      \\
		AANet\cite{AAnet}          & &45.40             & 8.05G    & 78.70          &94.40 \\
		Triplet Attention\cite{triplet_attention} & &44.56          &7.95G      & 78.03         & 93.85 \\
		EPSANet(Small)   &  & \textbf{38.90}   & \textbf{6.82G} & 78.43 & 94.11  \\
		EPSANet(Large)  &  & 49.59   & 8.97G  & \textbf{79.38}  & \textbf{94.58} \\
		\hline  
		
	\end{tabular}
	\label{table2:imagenet_prec}
\end{table}
\section{Experiments}
\subsection{Implementation Details}
\label{experiments_details}
For image classification tasks, we employ the widely used ResNet \cite{he2016deep} as the backbone model and perform experiments on the ImageNet \cite{szimagenet} dataset. The training configuration is set as the reference in \cite{res2net,he2016deep,hu2018senet}. Accordingly, the standard data augmentation scheme is implemented and the size of the input tensor is cropped to 224×224 by randomly horizontal fliping and normalization. The optimisation is performed by using the stochastic gradient descent (SGD) with weight decay of 1e-4, momentum as 0.9 and a minibatch size of 256. The Label-smoothing regularization \cite{labelsmooth_inceptionv2} is used with the coefficient value as 0.1 during training. The learning rate is initially set as 0.1 and is decreased by a factor of 10 after every 30 epochs for 120 epochs in total. For object detection tasks, the ResNet-50 along with FPN \cite{fpn} is used as the backbone model, we use three representative detectors, Faster RCNN \cite{ren2015faster}, Mask RCNN \cite{he2017mask} and RetinaNet \cite{retinanet} on the MS-COCO \cite{COCO} dataset. The default configuration setting is that the shorter side of the input image is resized to 800. The SGD is used with a weight decay of 1e-4, the momentum is 0.9, and the batch size is 2 per GPU within 12 epochs. The learning rate is set as 0.01 and is decreased by the factor of 10 at the 8th and 11th epochs, respectively. For instance segmentation tasks, we employ the main-stream detection system, Mask R-CNN \cite{he2017mask} and also in companion with FPN. The settings of training configuration and dataset are similar to that of the object detection. Finally, all detectors are implemented by the MMDetection toolkit \cite{mmdetection}, and all models are trained on 8 Titan RTX GPUs.

\subsection{Image Classification on ImageNet}
Table \ref{table2:imagenet_prec} shows the comparison results of our EPSANet with prior arts on ResNet with 50 and 101 layers. For the Top-1 accuracy, the proposed EPSANet-50(Small) achieves a margin of 2.29$\%$ higher over the ResNet-50, and using 11.7$\%$ fewer parameters and requires 12.1$\%$ lower computational cost. Moreover, with almost the same Top-1 accuracy, the EPSANet-50(Small) can save 54.2$\%$ parameter storage and 53.9$\%$ computation resources as compared to SENet-101. The EPSANet-101(Small) outperforms the original ResNet-101 and SENet101 by 1.6$\%$ and 0.81$\%$ in Top-1 accuracy, and saves about 12.7$\%$ parameter and 21.1$\%$ computational resources. With the similar Top-1 accuracy on ResNet-101, the computational cost is reduced about 12.7$\%$ by our EPSANet-101(Small) as compared to SRM, ECANet and AANet. What's more, our EPSANet-50(Large) shows the best performance in accuracy, achieving a considerable improvement compared with all the other attention models. Specifically, the EPSANet-50(large) outperforms the SENet, ECANet and FcaNet by about 1.93$\%$,1.16$\%$ and 0.12$\%$ in terms of Top-1 accuracy respectively. With the same number of parameters, our EPSANet-101(Large) achieves significant improvements by about 1.76$\%$ and 0.89$\%$ compared to the SENet101 and CBAM, respectively. In sum, the above results demonstrate that our PSA module has gain a very competitive performance with a much lower computational cost.
\begin{table}[H] \label{table:2}
	\centering
	\caption{Comparison of object detection results on COCO val2017}
	\resizebox{\textwidth}{45mm}{
		\setlength{\tabcolsep}{4pt}{
			\begin{tabular}{|l|l|c|c|c|c|c|c|c|c|}\hline
				Backbone & Detectors  &Parameters(M) &GFLOPs & AP & $AP_{50}$ & $AP_{75}$ & $AP_{S}$ & $AP_{M}$ & $AP_{L}$ \\ \hline
				ResNet-50\cite{he2016deep}  & \multirow{8}{10mm}{Faster-RCNN}  & 41.53 & 207.07 & 36.4 & 58.2 & 39.5 &21.8 & 40.0 & 46.2   \\ 
				SENet-50\cite{hu2018senet}          & & 44.02 & 207.18 & 37.7 & 60.1 & 40.9 & 22.9 & 41.9 & 48.2 \\
				ECANet-50\cite{20ecanet}   & & 41.53 & 207.18 & 38.0 & 60.6 & 40.9 & 23.4 & 42.1 & 48.0 \\   
				SANet-50\cite{SANet-attention}  & & 41.53 & 207.35 & 38.7 & 61.2 & 41.4 & 22.3 &42.5 &49.8 \\
				FcaNet-50\cite{2020FcaNet}  & & 44.02 & 215.63 & 39.0 & 61.1 & 42.3 & 23.7 &42.8 &49.6 \\
				EPSANet-50(Small)  &  &\textbf{38.56} &\textbf{197.07}   & 39.2 & 60.3  & 42.3  & 22.8  & 42.4  & 51.1 \\   
				EPSANet-50(Large)  &  &43.85  &219.64   &\textbf{40.9}  &\textbf{62.1} &\textbf{44.6}  &23.6  &\textbf{44.5} & \textbf{54.0}   \\
				
				\hline  
				ResNet-50\cite{he2016deep}  & \multirow{8}{10mm}{Mask-RCNN}  & 44.18 & 275.58  & 37.2  & 58.9  & 40.3 & 22.2  & 40.7  & 48.0   \\
				SENet-50\cite{hu2018senet}  &     & 46.67  & 275.69  & 38.7   & 60.9  & 42.1   & 23.4   & 42.7   & 50.0   \\
				Non-local\cite{Non-local_long}  &      & 46.50  & 288.70 & 38.0   & 59.8  & 41.0   & -     & -      & -    \\
				GCNet-50\cite{2019GCNet}                   &      & 46.90 & 279.60  & 39.4   & 61.6  & 42.4   & -     & -      & -    \\
				ECANet-50\cite{20ecanet}      &      & 44.18  & 275.69 & 39.0   & 61.3  & 42.1   & 24.2  & 42.8   & 49.9  \\
				SANet-50\cite{SANet-attention}  & & 44.18 & 275.86 & 39.4 & 61.5 & 42.6 & 23.4 &42.8 &51.1 \\
				FcaNet-50\cite{2020FcaNet} & &46.66 &261.93 &40.3 & 62.0 & 44.1 &25.2 & 43.9 & 52.0 \\
				EPSANet-50(Small)  &  & \textbf{41.20}   &\textbf{248.53}  & 40.0  & 60.9  & 43.3  & 22.3  & 43.2  & 52.8   \\   
				EPSANet-50(Large))  & & 46.50 & 271.10  &\textbf{41.4}  &\textbf{62.3}  &\textbf{45.3} &23.6  &\textbf{45.1}  &\textbf{54.6}  \\
				
				\hline
				ResNet-50\cite{he2016deep}  & \multirow{5}{10mm}{RetinaNet}  & 37.74 & 239.32 & 35.6  & 55.5  & 38.2   & 20.0   & 39.6   & 46.8      \\
				SENet-50\cite{hu2018senet}   & & 40.25 & 239.43 & 37.1 & 57.2 & 39.9 & 21.2 & 40.7 & 49.3  \\
				SANet-50\cite{SANet-attention}  &  & 37.74 & 239.60 & 37.5 & 58.5 & 39.7  & 21.3 & 41.2 & 45.9    \\
				EPSANet-50(Small)  &  & \textbf{34.78}  & \textbf{229.32}   &38.2  & 58.1  &40.6  &\textbf{21.5}  &41.5  &51.2   \\   
				EPSANet-50(Large))  &  &40.07  &251.89  &\textbf{39.6} &\textbf{59.4} &\textbf{42.3}  &21.2  &\textbf{43.4}  &\textbf{52.9}  \\
				
				\hline
	\end{tabular}}} 
	\label{detection_result}
\end{table}
\subsection{Object Detection on MS COCO}
As illustrated by Table \ref{detection_result}, our proposed models can achieve the best performance for the object detection task. Similar to the classification task on ImageNet, the proposed EPSANet-50(Small) outperforms the SENet-50 by a large margin with less parameters and lower computational cost. The EPSANet-50(Large) can achieve the best performance compared with the other attention methods. From the perspective of complexity (in term of parameters and FLOPs), the EPSANet-50(Small) offers a high competitive performance compared to the SENet50, i.e., by 1.5$\%$, 1.3$\%$, and 1.1$\%$, higher in bounding box $AP$ on the Faster-RCNN, Mask-RCNN, RetinaNet, respectively. What's more, as compared to the SENet50, the EPSANet-50(Small) can further reducing the number of parameters  to 87.5$\%$, 88.3$\%$ and 86.4$\%$ on Faster RCNN, Mask RCNN and RetinaNet, respectively. The EPSANet-50(Large) is able to boost the mean average precision by around 4$\%$ on the above three detectors as compared with the ResNet-50. It is worth noting that the most compelling performance improvement appears in the measurement of $AP_{L}$. With almost the same computational complexity, the $AP$ performance can be improved by 1.9$\%$ and 1.1$\%$ by our proposed EPSANet-50(Large) on both Faster-RCNN and Mask-RCNN detector, as compared to the FcaNet. The results demonstrate that the proposed EPSANet has good generalization ability and can be easily applied to other downstream tasks.
\begin{table}\centering
	\caption{Instance segmentation results of different attention networks by using the Mask R-CNN on COCO}
	\centering
	\begin{tabular}{|l|c|c|c|c|c|c|}\hline
		Network  & AP   & $AP_{50}$  & $AP_{75}$ & $AP_{S}$ & $AP_{M}$ & $AP_{L}$ \\ \hline
		ResNet-50 & 34.1 & 55.5 & 36.2 & 16.1 & 36.7  & 50.0 \\ 
		SENet-50 & 35.4  & 57.4  & 37.8  & 17.1 & 38.6  & 51.8 \\
		ResNet-50 + 1 NL-block\cite{Non-local_long} & 34.7 & 56.7 & 36.6 & -  & - & - \\ 
		GCNet & 35.7 & 58.4 & 37.6 & - & - & - \\ 
		ECANet & 35.6  & 58.1 & 37.7 & 17.6 & 39.0 & 51.8 \\ 
		FcaNet & 36.2  & 58.6 & 38.1 & - & - & - \\
		SANet  & 36.1 & 58.7 & 38.2 & 19.4 &39.4 &49.0 \\
		EPSANet-50(Small)& 35.9  & 57.7  & 38.1 & 18.5  & 38.8 & 49.2 \\
		EPSANet-50(Large)& \textbf{37.1}  &\textbf{59.0}  & \textbf{39.5} & \textbf{19.6} & \textbf{40.4}& 50.4 \\
		\hline 
	\end{tabular}
	\label{instance_result}
\end{table}

\subsection{ Instance Segmentation on MS COCO}
For instance segmentation, our experiments are implemented by using the Mask R-CNN on MS COCO dataset. As illustrated by Table \ref{instance_result}, our proposed PSA module outperforms the other channel attention methods by a considerably larger margin. Specifically, our EPSANet-50(Large) surpass the FcaNet which can offer the best performance in existing methods, by about 0.9$\%$ , 0.4$\%$ and 1.4$\%$ on $AP$, $AP_{50}$ and $AP_{75}$ respectively. These results verified the effectiveness of our proposed PSA module.
\begin{table}
	\centering
	\caption{Accuracy performance with the change of group size}
	\begin{tabular}{|l|l|c|c|}\hline
		Kernel size  & Group size     & Top-1 Acc($\%$)   &Top-5 Acc($\%$) \\
		\hline
		(3,5,7,9)   &(4,8,16,16)     &77.25  &93.40 \\ \hline
		(3,5,7,9)   &(16,16,16,16)   &77.24  &93.47 \\ \hline
		(3,5,7,9)   &(1,4,8,16)      &\textbf{77.49}  &\textbf{93.54} \\ \hline
	\end{tabular}
	\label{ablation}
\end{table}
\subsection{Ablation Study}
\textbf{Kernel and Group sizes}
As shown by Table \ref{ablation}, we adjust the group size to verify the effectiveness of our network on the ImageNet \cite{szimagenet} dataset. A significant increase in the amount of parameters will be resulted by increasing kernel sizes in parallel. In order to exploit the location information of multi-scale in spatial domain without increasing the computational cost, we apply the group convolution independently for each feature map with different scale. By properly adjusting the group size, the number of parameters and the computational cost can be reduced by a factor which equals to the number of sub-groups. As shown by Table \ref{ablation}, a good balance between the performance and model complexity can be achieved by our proposed EPSANet.

\section{Conclusion}
In this paper, an effective and lightweight plug-and-play attention module named Pyramid Squeeze Attention(PSA) is proposed. The proposed PSA module can fully extract the multi-scale spatial information and the important features across dimensions in the channel attention vector. The proposed Efficient Pyramid Squeeze Attention(EPSA) block can improve the multi-scale representation ability at a more granular level and develop a long-range channel dependency. The proposed EPSANet can effectively intergrate multi-scale contextual features and image-level categorical information. With extensive qualitative and quantitative experiments, it is verified that the proposed EPSANet can achieve state-of-the-art performance across image classification, object detection and instance segmentation compared with the other conventional channel attention methods. We will investigate the effects of adding PSA module to more lightweight CNN architectures as our future work.

\section*{Acknowledgement}
This work was supported by the National Natural Science Foundation of China under grants 61972265 and 11871348, by the Natural Science Foundation of Guangdong Province of China under grant 2020B1515310008, by the Educational Commission of Guangdong Province of China under grant 2019KZDZX1007, the Pazhou Lab, Guangzhou, China.

\bibliographystyle{unsrt}
\bibliography{main}

\end{document}